\documentclass{article}

\usepackage{microtype}
\usepackage{graphicx}
\usepackage{subfigure}
\usepackage{amssymb}
\usepackage{amsmath}
\usepackage{amsthm}
\usepackage{placeins}
\usepackage{enumitem}

\usepackage{booktabs}
\usepackage{dblfloatfix}

\usepackage{hyperref}
\usepackage[nolist]{acronym}

\renewcommand{\vec}[1]{\mathbf{#1}}

\begin{acronym}
\acro{RNN}{Recurrent Neural Network}
\acro{GRU}{gated recurrent unit}
\acro{LSTM}{long short-term memory}
\acro{MFCC}{Mel-frequency cepstral coefficient}
\acro{PEM}{per-epoch noise mixing}
\acro{SNR}{signal-to-noise ratio}
\acro{ASR}{automatic speech recognition}
\acro{CNN}{Convolutional Neural Network}
\acro{DNN}{Deep Neural Network}
\acro{ACCAN}{accordion annealing}
\acro{WSJ}{Wall Street Journal}
\acro{CTC}{Connectionist Temporal Classification}
\acro{WFST}{Weighted Finite State Transducer}
\acro{WER}{word error rate}
\acro{CER}{character error rate}
\acro{LVCSR}{large-vocabulary continuous speech recognition}
\acro{ROI}{range of interest}
\acro{STAN}{sensor transformation attention network}
\acro{SER}{sequence error rate}
\acro{HMM}{Hidden Markov Model}
\acro{LER}{label error rate}
\acro{NN}{Neural Networks}
\acro{MLP}{Multilayer Perceptron }
\acro{ReLU}{Rectified Linear Unit}
\acro{PCC}{Pearson correlation coefficient}
\acro{LPN}{Lightweight Probabilistic Deep Network}
\end{acronym}

\usepackage[accepted]{icml2019}

\icmltitlerunning{Deep Approximate Shapley Propagation}

\begin{document}

\twocolumn[
\icmltitle{Explaining Deep Neural Networks with a Polynomial Time Algorithm for Shapley Values Approximation}

\begin{icmlauthorlist}
\icmlauthor{Marco Ancona}{eth}
\icmlauthor{Cengiz {\"O}ztireli}{drs}
\icmlauthor{Markus Gross}{eth,drs}
\end{icmlauthorlist}

\icmlaffiliation{eth}{Department of Computer Science, ETH Zurich, Switzerland}
\icmlaffiliation{drs}{Disney Research Zurich, Switzerland}

\icmlcorrespondingauthor{Marco Ancona}{marco.ancona@inf.ethz.ch}

\icmlkeywords{Neural Networks, Attribution methods, Shapley values, Explainable AI, XAI}

\vskip 0.3in
]

%\printAffiliations{} % need to use arxive style 
\printAffiliationsAndNotice{}

\begin{abstract}
The problem of explaining the behavior of deep neural networks has recently gained a lot of attention. While several attribution methods have been proposed, most come without strong theoretical foundations, which raises questions about their reliability.
On the other hand, the literature on cooperative game theory suggests Shapley values as a unique way of assigning relevance scores such that certain desirable properties are satisfied. 
Unfortunately, the exact evaluation of Shapley values is prohibitively expensive, exponential in the number of input features. 
In this work, by leveraging recent results on uncertainty propagation, we propose a novel, polynomial-time approximation of Shapley values in deep neural networks. 
We show that our method produces significantly better approximations of Shapley values than existing state-of-the-art attribution methods.
\end{abstract}

\section{Introduction}\label{sec:intro}
\acp{DNN} have demonstrated enormous potential in solving a variety of problems, growing the sophistication and impact of machine learning. 
On the other hand, while machine learning models are being employed on an increasing number of fields, the black-box nature of \acp{DNN} is still a barrier to the adoption of these systems for those tasks where interpretability is a requirement.
Recently, European regulators introduced the legal notion of a \textit{right to explanation} \cite{goodman2016european}, demanding transparency for any automated decision having a deep impact on the life of the people involved, which affects the adoption of black-box models like \acp{DNN} on some domains.

Since the notion of interpretability is complex, multi-faceted and not yet fully defined \cite{finale2017towards, lipton2016mythos}, several works have focused on investigating methods for \textit{local interpretability}. Contrarily to global interpretability, which aims at explaining the general model behavior, local interpretability scope is restricted at explaining a particular decision for a given model and input instance \cite{finale2017towards}.

In the realm of local interpretability, \textit{attribution methods} have received particular attention in the last years \cite{ras2018explanation}. 
Consider a model that takes an N-dimensional input $\vec{x} = [x_1,...,x_N] \in \mathbb{R}^N$ and produces a C-dimensional output $\vec{f(x)} = [{f}_1(\vec{x}),...,{f}_C(\vec{x})] \in \mathbb{R}^C$, like a \ac{DNN} with $C$ output neurons. 
Depending on the application, the input features $x_1, ..., x_N$ can have a different nature, like pixels for images or components of a multi-dimensional word vector representation for natural language processing.
Similarly, each output of the network can represent either a numerical predicted quantity (regression task) or the probability of a corresponding class (classification task).

Formally, attribution methods aim at producing explanations by assigning a scalar \textit{attribution} value, sometimes also called ``relevance'' \cite{bach2015pixel} or ``contribution'' \cite{shrikumar17a}, to each input feature of a network for a given input sample.
In particular, for a single target output indexed with $c$, the goal of an attribution method is to determine the contribution $\vec{R}^c(\vec{x};f_c) = [R_1^c(\vec{x};f_c),...,R_N^c(\vec{x};f_c)] \in \mathbb{R}^N$ of each input feature $x_i$ to the output ${f}_c(\vec{x})$. For clarity of notation, in the remainder of the paper, we will simply use $\vec{R}^c$ whenever possible. For a classification task, the target output can be chosen to be the one associated with the highest output probability (to understand which part of the input was mostly relevant for the prediction) or the one associated with a different class (to assess whether the input contains evidence that supports or rejects a different class).
Over the last decade, several attribution methods have been specifically developed for neural networks \citep{simonyan2013deep, zeiler2014visualizing, springenberg2014striving,  bach2015pixel,ribeiro2016why,selvarajuDVCPB16, shrikumar17a, sundararajan17a, montavon2017explaining, zintgraf2017visualizing, lundberg2017unified, kindermans2018learning}.

When an explanation is generated, it should be a natural requirement to have a clear understanding of how the explanation method works, a condition necessary to make sure the explanation itself is a reliable and unbiased representation of the network behavior. In fact, some recent works showed that attribution methods can actually produce unreliable or misleading results, despite these being visually appealing to humans \cite{kindermans2017reliability,ghorbani2017interpretation,adebayo2018sanity, nie2018theoretical}. This problem is fueled by the limited theoretical understanding of some of these methods and lack of reliable quantitative metrics to evaluate explanations in the absence of a ground-truth \cite{adebayo2018sanity}. To overcome these limitations, some authors have recently endorsed an \textit{axiomatic approach} \cite{Sun:2011:AAM:1993574.1993601, sundararajan17a, montavon2017explaining, lundberg2017unified, kindermans2017reliability}. In this context, an axiom is a self-evident property of the attribution method that should be satisfied for any explanation generated by the method itself. By leveraging these properties, attribution methods with stronger theoretical guarantees can be designed \cite{sundararajan17a}.

Along with this research direction, the literature on cooperative game theory suggests Shapley values \cite{shapley1953value} as a unique way of assigning attributions such that certain desirable axioms are satisfied. Shapley values are a classic game theory solution for the distribution of credits to players participating in a cooperative game. Notably, the unique set of properties of Shapley values, discussed in the next section, seems to fit naturally in the setting of attributions as well \cite{strumbelj2010efficient, Sun:2011:AAM:1993574.1993601, datta2016algorithmic,  lundberg2017unified}.

% %
On the other hand, computing the exact Shapley value is, in general, NP-hard \cite{matsui2001np} and only feasible for less than 20-25 players (i.e. input features for our case). Some previous works proposed sampling-based methods to approximate Shapley values \cite{castro2009, strumbelj2010efficient, datta2016algorithmic}.  While these methods are unbiased estimators, as the number of input features grows they require thousands of model evaluations, which is expensive for \acp{DNN}. KernelSHAP \cite{lundberg2017unified} combines sampling with lasso regression to reduce the number of samples but it also introduces a bias from the regularizer. Other fast approximations designed for \acp{DNN} exist but these are based on the strong assumptions of model linearity \cite{lundberg2017unified}. At the best of our knowledge, there is no exhaustive evaluation of the empirical accuracy of these methods as Shapley value approximators in \acp{DNN}.

The contribution of this work is threefold: 1) endorsing an axiomatic approach, we compare Shapley values to existing state-of-the-art attribution methods, motivating the use of the former to explain non-linear models; 2) we formulate a novel, polynomial-time approximation of Shapley values specifically designed for \acp{DNN}; 3) we assess empirically the approximation power of our algorithm compared to other attribution methods on three datasets and architectures.

\section{Attribution Methods}
Attribution methods can be partitioned into two broad categories: \textit{backpropagation-based methods} and \textit{perturbation-based methods} \cite{ancona2018towards}. In this section, we define Shapley values and briefly review some popular attribution methods that will be later used in our experimental comparison. We focus on methods that can be applied to a variety of architectures and that are not restricted to some particular input type (eg. images).

\subsection{Backpropagation-based methods}
Methods in this category compute attributions running only one or a few backward passes through the network. Many such methods have been proposed in the last years. The prototypical example is Saliency maps \cite{simonyan2013deep} where attributions are computed as the absolute gradient of the target output with respect to each input feature.
 Gradient $\times$ Input \cite{shrikumar2016not} introduces an element-wise multiplication between the input and the (signed) gradient, producing more sharp results:
 \begin{equation}
     R_i^c = x_i \cdot \frac{\partial {f}_c(\vec{x})}{\partial x_i}
 \end{equation}
While the gradient provides information about which features can be locally perturbed the least in order for the output to change the most, applied to a highly non-linear function only provides local information and it does not help to compute the marginal contribution of a feature. To overcome this limitation, other methods have been proposed such as Layer-wise Relevance Propagation (LRP) in several variants \cite{bach2015pixel, montavon2017explaining}, DeepLIFT (in its two variants, Rescale and RevealCancel) \cite{shrikumar2016not, shrikumar17a}, and Integrated Gradients \cite{sundararajan17a}.
Compared to Gradient $\times$ Input, these methods are characterized by different propagation rules for non-linear operations in the network than the instant gradient. In the case of Integrated Gradients, DeepLIFT Rescale, and $\epsilon$-LRP, the propagation rule can be seen as computing a form of \textit{average gradient} \cite{ancona2018towards}. In the other cases, the propagation rule is designed on specific heuristics.

\subsection{Perturbation-based methods}
This category includes methods that estimate the contribution of input features by removing or perturbing them and measuring the variation of the network target output as a consequence of this operation \cite{ribeiro2016why, zintgraf2017visualizing, fong2017interpretable}. The simplest perturbation method computes attributions by setting each feature sequentially to zero \cite{zeiler2014visualizing}:
 \begin{equation}
     R_i^c = {f}_c(\vec{x}) - {f}_c(\vec{x} \setminus \{x_i\})
 \end{equation}
 In the remainder of the paper, we will refer to this method simply as ``Occlusion''. 
 
 \textbf{Remark} Notice that the procedure of replacing features with a zero value implicitly defines a \textit{baseline} that can be used to indicate features that are toggled off. Many attribution methods require the definition of a baseline to indicate the absence of information. This is further discussed in Appendix C of the supplementary material.
 
Most often perturbation-based methods are simple to implement but very slow, as several evaluations of the network are necessary. Additionally, the number of features perturbed at each iteration and the choice of the perturbation itself are hyper-parameters of these methods that can heavily affect the resulting explanations, making it difficult to interpret the results \cite{ancona2018towards}.

 \subsection{Shapley values}
 Shapley values can be considered a particular example of perturbation-based methods where no hyper-parameters, except the baseline, are required.
 
 Consider a set of $N$ players $P$ and a function $\vec{\hat{f}}$ that maps each subset $S \subseteq P$ of players to real numbers, modeling the outcome of a game when players in $S$ participate in it. The Shapley value is one way to quantify the total contribution of each player to the result $\vec{\hat{f}}(P)$ of the game when all players participate. For a given player $i$, it can be computed as follows:
\begin{equation}
    R_i = \sum_{S \subseteq P \setminus \{i\}} \frac{|S|!(|P| - |S| - 1)!}{|P|!} [\vec{\hat{f}}(S \cup \{i\}) - \vec{\hat{f}}(S)]
\label{eq:shapley}
\end{equation}
The Shapley value for player $i$ defined above can be interpreted as the \textit{average marginal contribution} of player $i$ to all possible coalitions $S$ that can be formed without it.

While $\vec{\hat{f}}$ is a set function, the definition can be adapted for a neural network function $\vec{f}$ by defining a baseline as discussed above. Then we can replace $\vec{\hat{f}}(S)$ in Eq.\ \ref{eq:shapley} with $\vec{f}(\vec{x}_S)$, where $\vec{x}_S$ indicates the original input vector $\vec{x}$ where all features not in $S$ are replaced with the baseline value. 

It has been observed \cite{lundberg2017unified} that attributions based on Shapley values better agree with the human intuition empirically. Unfortunately, computing Shapley values exactly requires us to evaluate all $2^N$ possible feature subsets as in Eq.\ \ref{eq:shapley}. This is clearly prohibitive for more than a couple of dozens of variables. 

For certain simple functions $\vec{f}$, it is possible to compute Shapley values exactly in polynomial time. For example, the Shapley values of the inputs to a \textit{max-pooling} layer can be computed with $\mathcal{O}(N^2)$ function evaluations \cite{lundberg2017unified}. When this is not possible, Shapley values sampling \cite{castro2009} can be used to estimate approximate Shapley values based on sampling of Eq.\ \ref{eq:shapley}. For specific games, other approximate methods have been developed. In particular, \cite{fatima2008linear} proposed a polynomial time approximation for \textit{weighted voting games} \cite{osborne1994course} based on the idea that, instead of enumerating all coalitions, it might suffice to estimate their expected contributions. The same idea is used as the first step of our approximation algorithm for deep neural networks in Sec.\ \ref{sec:deepapproxshapleyprop}. Before going into the details of the derivations, we present a review of theoretical properties of Shapley values and other attribution methods.

\section{Axiomatic comparison of attribution methods}

We start our theoretical analysis by observing the following connection between existing attribution methods and Shapley values for linear models.

\newtheorem{prop}{Proposition}
\begin{prop}\label{prop:lrp}Occlusion, Gradient $\times$ Input, Integrated Gradients and DeepLIFT produce exact Shapley values when applied to a linear model and a zero baseline is used.
\end{prop}

The proof (provided in Appendix A of the supplementary material) comes directly from the observation that all these methods are equivalent for linear models \cite{ancona2018towards}.
For non-linear models, instead, these methods produce different attributions. DeepLIFT RevealCancel \cite{shrikumar17a} and its variant DeepSHAP \cite{lundberg2017unified} approximate Shapley values at each layer independently and use the chain rule to compose these into attributions. 
Unfortunately, the chain rule does not hold in general for Shapley values.
Integrated Gradients \cite{sundararajan17a}, on the other hand, can be shown to compute Aumann-Shapley values \cite{aumann1974values}, an extension of Shapley values for infinite games where each individual player has a negligible contribution. While this method has better theoretical properties, it is not yet clear how well these assumptions apply to real scenarios where the number of input features is limited and the individual features may have a significant impact.

In order to better understand the relations among methods and their advantages, it is essential to evaluate them with respect to desirable theoretical properties. We thus compare different attribution methods according to the following axioms already established in the literature:
\textit{(1) conservation},
\textit{(2) null player},
\textit{(3) symmetry},
\textit{(4) linearity},
\textit{(5) continuity} and 
\textit{(6) implementation invariance}.

\textbf{Axiom 1: Completeness.} An attribution method satisfies \textit{completeness} when attributions sum up to the difference between the value of the function evaluated at the input, and the value of the function evaluated at the baseline, i.e. $\sum_{i=1}^n R_i^c = \vec{f}_c(\vec{x}) - \vec{f}_c(\vec{0})$. This property, also called \textit{efficiency} \cite{shapley1953value}, \textit{summation to delta} \cite{shrikumar17a} or \textit{conservation} \cite{montavon2017explaining},  has been recognized by previous works as desirable to ensure the attribution method is comprehensive in its accounting.

\textbf{Axiom 2: Null player.} If the function implemented by a \ac{DNN} does not depend on some variable, then the attribution to that variable should always be zero.

\textbf{Axiom 3: Symmetry.} If the function implemented by a \ac{DNN} depends on two variables $x_1$ and $x_2$ but not on their order (i.e. $f(x_1, x_2) = f(x_2, x_1)$), then the attribution assigned to these variables should be the same every time the input and the baseline provides the same values for these variables. This axiom, also called \textit{anonymity} \cite{Sun:2011:AAM:1993574.1993601}, is arguably a desirable property for any attribution method: if two variables play the exact same role
in the \ac{DNN}, they should receive the same attribution.

\textbf{Axiom 4: Linearity.} If the function $f$ implemented by a \ac{DNN} can be seen as a linear combination of the functions of two sub-networks (i.e. $f = a \times f_1 + b\times f_2$), then any attribution should also be a linear combination, with the same weights, of the attributions computed on the sub-networks, i.e. $R_i^c(\vec{x}|f) = a\times R_i^c(\vec{x}|f_1) + b \times R_i^c(\vec{x}|f_2)$, where $R_i^c(x|f)$ denotes the attributions for the DNN that implements the function $f$. 
Intuitively, this is justified by the need for preserving linearities within the network \cite{sundararajan17a}.

\textbf{Axiom 5: Continuity} Attributions generated for two nearly identical inputs should be nearly identical, i.e. $R_i^c(\vec{x}) \approx R_i^c(\vec{x}+\boldsymbol{\epsilon})$. This axiom assumes the attribution is generated for a continuous prediction function $f(\vec{x})$, which is always the case for \acp{DNN} built by combining continuous functions. In this case, the prediction for two nearly identical inputs is also nearly identical, which motivates a nearly identical explanation \cite{montavon2017explaining}.

\textbf{Axiom 6: Implementation Invariance.} Two networks are said to be \textit{functionally equivalent} if their outputs are equal for all inputs, despite having (possibly very) different implementations \cite{sundararajan17a}. Attribution methods should produce identical results when applied to any functionally equivalent network provided with the same input. The importance of this property relies on the observation that any explanation should only depend on the input and the function implemented by the \ac{DNN}, not its implementation. This property is satisfied by all perturbation-based methods (that only relies on the function evaluation) as well as by all backpropagation-based methods relying purely on the gradient (which is itself implementation-invariant). This axiom was proposed in a previous work \cite{sundararajan17a} where it is shown that it can be violated when other propagation rules are used, like in DeepLIFT and LRP.

%\vspace{0.2in}
It is trivial to see that Gradient $\times$ Input and Occlusion do not satisfy Completeness, while it has been showed that LRP and DeepLIFT do not satisfy Implementation Invariance \cite{sundararajan17a}. Instead, the following notable result can be shown for Shapley values.

\begin{prop}\label{prop:shapley}Shapley values is the only possible attribution method that satisfies Axioms 1-5.
\end{prop}
\vskip -0.01in
The proposition is a direct consequence of previous results \cite{Sun:2011:AAM:1993574.1993601} and the proof is provided in Appendix B of the supplementary material.
Clearly, being a perturbation-based method, Shapley values satisfies Axiom 6 as well. This unique set of sought after properties motivates the use of Shapley values for attributions.

\section{Deep Approximate Shapley Propagation}
\label{sec:deepapproxshapleyprop}
Despite the theoretical guarantees, Shapley values have one big flaw: computing them is NP-hard, as elaborated on in the previous sections. In this section, we introduce Deep Approximate Shapley Propagation (DASP), a perturbation-based method that can reliably approximate Shapley values in \acp{DNN} with a polynomial number of perturbation steps.

Consider a feed-forward neural network, composed of layers equipped with a non-linear function $\sigma$, each performing 
\begin{equation}\label{eq:layer}
    \vec{f}^{(i)}(\vec{x}^{(i-1)}) = \sigma(\vec{W}^{(i)} \vec{x}^{(i-1)}  + \vec{b}^{(i)})
\end{equation}
where $i$ indicates a layer index, $1 \leq i \leq l$.

We are interested in computing the Shapley values with respect to one output unit of a neural network whose overall function $\vec{f}(\vec{x}) = (\vec{f}^{(1)} \circ \vec{f}^{(2)} \circ ... \circ \vec{f}^{(l)}) (\vec{x})$ is the composition of several of such layers.

Our approximation is based on the following intuition. According to the definition given by Eq.\ \ref{eq:shapley}, the Shapley value of an input feature is given by its marginal contribution to all possible $2^{N-1}$ coalitions that can be made out of the remaining features. Since we are interested in an \textit{average} value, we can compute the expected contribution to a random coalition instead of enumerating each of them. In particular, we consider the distribution of coalitions of size $k$, for each $0 \leq k \leq N - 1$, that do not include the feature $x_i$, and compute the expected contribution of that feature with respect to these distributions.
The average of all these marginal contributions gives the feature's approximate Shapley value \cite{fatima2008linear}:

\begin{equation}
    \label{eq:fatima}
    \mathbb{E}\big[R_i^c\big] = \frac{1}{N} \sum_{k=0}^{N-1} \mathbb{E}_k \big[R_{i,k}^c \big],
\end{equation}
where the expectations $\mathbb{E}_k$ are over the distribution of sets of size $k$, and $R_{i,k}^c$ denotes the contribution of feature $x_i$ to any random coalition of size $k$.
More explicitly, we can write the expected contribution of a feature $x_i$ for a given coalition size as the expected target output difference with and without it, i.e.
\vskip -0.2in
\begin{equation}
\begin{split}
    \mathbb{E}_k\big[R_{i,k}^c\big] 
    &= \mathop{\mathbb{E}}_{\substack{S \subseteq P \setminus \{i\} \\ |S|=k}} [{f}_c(\vec{x}_{S \cup \{i\}})] - \mathop{\mathbb{E}}_{\substack{S \subseteq P \setminus \{i\} \\ |S|=k}} [{f}_c(\vec{x}_S)].
\end{split}
\label{eq:expectedValues}
\end{equation}

\vskip -0.2in
The proof of Eq.\ \ref{eq:fatima} is based on the observation that there exists the same number of coalitions (where order matters) of size $k$ for all values of $k$. We refer the reader to \cite{fatima2008linear} for the complete derivation.

So the main problem is approximating these expected values for all coalitions of size $0 \leq k \leq N-1$. As we will elaborate on in the next sections, these expected values can be computed and propagated along with variances from layer to layer in a DNN. Such a propagation can be achieved by transforming the architecture of a given DNN to replace the point activations at all layers by probability distributions. 
This problem has been previously studied in the scope of uncertainty propagation in \acp{DNN} \cite{abdelaziz2015uncertainty, gast2018lightweight, thiagarajan2018understand}, where the goal is to analyze how the uncertainty of the input data or of the network parameters propagates through the linear and non-linear operations up to the output layer. We adapt such a framework for our use case by \textit{i}) considering the network parameters fixed and hence no source of uncertainty, and \textit{ii}) introducing input uncertainty based on sampling of input coalitions. Then, we employ a Lightweight Probabilistic Network \cite{gast2018lightweight}, and show that the expected values can be propagated in a practical fashion through the entire network. 

\subsection{Input distribution from random coalitions}
\label{sec:inputDistFromRndCoal}
Each coalition of size $k$ is represented with a vector $\vec{x}_S$ of inputs, where the set $S$ of $k$ components of the vector contain their actual values, and the others contain the baseline value of zero. All such vectors with $k$ non-zero elements can be thought of as drawn from an underlying distribution of a random variable $\vec{X}_k$. 

The first operation in a typical DNN is a weighted sum of the inputs. Each hidden unit thus takes a weighted sum of the network input $z_{j} = \sum_{i=1}^N x_i \cdot w_{ij}$. As $\vec{X}_k$ is a random variable, a weighted sum of its components is also a random variable. It can be shown \cite{von1972sampling} that, under mild assumptions, the distribution of $Z_j$ is given by
\begin{equation}
    Z_j \sim \mathcal{N}\Bigg(k\mu_j, k \sigma_j^2  \frac{N-k}{N-1}\Bigg),
\end{equation}
where we set
\begin{equation}
    \mu_j = \frac{1}{N} \sum_{i=1}^{N} x_i \cdot w_{ij}, \quad \sigma^2_j = \frac{1}{N} \sum_{i=1}^{N} (x_i \cdot w_{ij})^2 - \mu_j^2,
\end{equation}
with the sums computed over all components $x_i$ of the input vector $\vec{x}$. We provide a study of the accuracy of this form of the distribution by comparing it to the empirical counterpart in Appendix D of the supplementary material.

The random vector $\vec{Z}$ with the components $Z_j$ is thus distributed with an isotropic Gaussian of the means $\mu_j$ and variances $\sigma^2_j$. Note that if we assume that this weighted sum is the only layer in the network, then the means here already provide the expected values we need to compute Shapley values. As this random variable is fed into the further operations of the later layers in a DNN, the distribution and hence the means will change. The next task is thus to derive how the distribution of $\vec{Z}$ will change as it is fed through.

\subsection{Distribution propagation}
Given the probabilistic input $\vec{Z}$ at the first hidden layer, we propagate this distribution to the target unit by employing \acp{LPN} \cite{gast2018lightweight}. \ac{LPN} is an architecture for uncertainty propagation through a feed-forward \ac{DNN} where
each input sample is modeled using an independent univariate Gaussian distribution. Contrary to other Bayesian formulations, the model parameters are considered deterministic, as in our case.
The propagation of uncertainties is carried out using \textit{assumed density
filtering} \cite{Boyen:1998:TIC:2074094.2074099, gast2018lightweight}, where each layer is implemented as filtering of the input distribution to obtain a transformed Gaussian distribution with diagonal covariance.  While it was originally developed to propagate the intrinsic uncertainty of the input, LPN can be easily adapted to our scenario, where the uncertainty given by random coalitions induces a probability distribution for the input of the first hidden layer. On the other hand, notice that DASP is not coupled to any particular probabilistic framework and can extend to general architectures (eg. RNNs) given a probabilistic framework that supports them.

% Mathematically, the joint density of all activations is given by
% \begin{equation}
%     p(\vec{X}^{(0:l)}) = p(\vec{X}^{(0)}) \prod_{i=1}^l (p(\vec{X}^{(i)} | \vec{X}^{(i-1)}),
% \end{equation}
% where $l$ denotes the number of layers, $\vec{X}$ are activations and $\vec{X}^{(0)} = \vec{Z}$ is a probabilistic input.
% %
% Given the independent Gaussian assumption, we can write
% \begin{equation}
%     p(\vec{X}^{(i)}) = \prod_j \mathcal{N}(X_j | \mu_j^{(i)}, (\sigma^{(i)}_j)^2)
% \end{equation}
% where $(\mu_j^{(i)}, \sigma^{(i)}_j)$ are the mean and standard deviation of the activation of unit $j$ at layer $i$. 

In order to propagate the activation distribution through linear and non-linear operations, LPN converts any layer with point activations into
an uncertainty propagation layer by matching first and second-order central moments, i.e.
\begin{subequations}
\begin{eqnarray}
\boldsymbol{\mu}_{\vec{X}}^{(i)} = \mathbb{E}_{\vec{X}^{(i-1)}} \big[ \vec{f}^{(i)}(\vec{x}^{(i-1)}) \big] \\
\boldsymbol{\sigma^2}_{\vec{X}}^{(i)} =
\mathbb{V}_{\vec{X}^{(i-1)}} \big[ \vec{f}^{(i)}(\vec{x}^{(i-1)}) \big],
\end{eqnarray}
\end{subequations}

where $\mathbb{E}[\cdot]$ and $\mathbb{V}[\cdot]$ denote expectation and variance. This procedure can be proven to perform a greedy (layer-wise) optimization of the KL-divergence of $p(\vec{x}^{(0:l)})$ towards the actual distribution \cite{Minka:2001:FAA:935427}.
As a result, our original network function $\vec{f}(\vec{x})$ is transformed into a probabilistic network function $\hat{\vec{f}}(\boldsymbol{\mu}, \boldsymbol{\sigma^2}) = [\hat{f}_1(\boldsymbol{\mu}, \boldsymbol{\sigma^2}), ..., \hat{f}_C(\boldsymbol{\mu}, \boldsymbol{\sigma^2})]$ that takes as inputs the first and second moments of a (Gaussian) distribution and returns the parameters of the distribution at the output layer, after the original distribution is sequentially filtered by all hidden layers.
Moment matching for several commonly employed layers and operations in neural network architectures can be derived in closed-form:

\textbf{Affine transformation}. A fully-connected layer with weights $\vec{W}$ and bias $\vec{b}$ takes the input $\vec{Z} \sim \mathcal{N}(\boldsymbol{\mu}, \boldsymbol{\sigma^2})$ and filters it by applying $\vec{f}(\vec{Z}) = \vec{W}\vec{Z} + \vec{b}$. The output is a new Gaussian with mean and variance:
\begin{equation}
    \boldsymbol{\mu}_{lin} = \vec{W}\boldsymbol{\mu} + \vec{b}; \qquad \boldsymbol{\sigma^2}_{lin} = \vec{W}^2 \boldsymbol{\sigma^2},
\end{equation}
where by $\vec{W}^2$ we indicate the element-wise square. This notation will be used for any matrix or vector in the remainder of the paper. With minor adaptations, this also holds for the convolution and the mean pooling, which are linear layers. Notice that mean and variance do not interact with each other on linear operations.

\textbf{ReLU activation}. The output of a ReLU activation that receives a Gaussian input $\vec{Z} \sim \mathcal{N}(\boldsymbol{\mu}, \boldsymbol{\sigma}^2)$ is a \textit{rectified Gaussian distribution} \cite{socci1998rectified} whose mean and variance can be derived analytically \cite{Frey:1999:VLN:307835.307858}. For details, see Appendix E of the supplementary material.

% \textbf{Mean Pooling}. Mean pooling is a special case of a linear operator that takes $n$ random variables $Z_1, ... Z_n$, and filters them to $\frac{1}{n} \sum_{i=1}^n Z_i$. The output moments will be
% \begin{equation}
%     {\mu}_{mean} = \frac{1}{n}\sum_{i=1}^n {\mu_i}; \qquad {\sigma}^2_{mean} = \frac{1}{n^2}\sum_{i=1}^n {\sigma^2_i}
% \end{equation}

\textbf{Max Pooling}. Max pooling can be seen as returning the maximum response of $n$ random variables $Z_1, ... Z_n$. For two independent inputs $A \sim \mathcal{N}({\mu_A}, {\sigma^2_A})$, $B \sim \mathcal{N}({\mu_B}, {\sigma^2_B})$, the maximum is not normally distributed anymore. Nevertheless, it has been shown that the univariate normal is an effective approximation \cite{gast2018lightweight} and the first and second moments, provided also in Appendix E, can be derived analytically \cite{Jacobs2000}.

\subsection{Computing approximate Shapley values}
Having transformed the distribution of inputs from coalitions to output activation uncertainties, we can finally compute approximate Shapley values. Algorithm \ref{alg:deepshaply} describes the procedure in pseudo-code. For the sake of clarity, we have listed all the operations of the first weighted sum as explained in Sec.\ \ref{sec:inputDistFromRndCoal} explicitly, and wrapped all other operations in subsequent layers into $\hat{f}_c$.

\begin{algorithm}[tbh!]
   \caption{Deep Shapley algorithm (dense layers only)}
   \label{alg:deepshaply}
\begin{algorithmic}[1]
   \STATE {\bfseries Input:} input $\vec{x}$, coalitions sizes $k_1, ..., k_K$, first layer weights $\vec{W}$, LPN without first linear layer $\hat{f}_c$
   \STATE Initialize result vector $\vec{R}^c$ at zero
   \FOR{$i=1, ..., N$}
   \FOR{$k=k_1, ..., k_K$}
   
   \STATE $\vec{\bar{x}} = \vec{x}$

   \STATE $\vec{\bar{x}}[i] = 0$
   
   \STATE // Compute statistics of features excluding $i$
    \STATE $\boldsymbol{\mu} = \frac{1}{N-1}(\vec{W}\vec{\bar{x}})$
    
   \STATE $\boldsymbol{\sigma^2} =  \frac{1}{N-1}(\vec{W}^2\vec{\bar{x}}^2 ) - \boldsymbol{\mu}^2$
   
   \STATE // Compensate for current coalition size
       \STATE $\boldsymbol{\mu} = k \boldsymbol{\mu}$
   \STATE $\boldsymbol{\sigma^2} = k \frac{N-k}{N-1}  \boldsymbol{\sigma^2}$
   
   \STATE // Compute bias introduced by $i$
    \STATE $\boldsymbol{\bar{\mu}} = \boldsymbol{\mu} + \vec{W}[:,i]\vec{x}[i]$
      
   \STATE // Propagate distributions up to the output layer 
   \STATE ${\mu}^{(l)}, {\sigma^{2}}^{(l)} = \hat{f}_c(\boldsymbol{\mu}, \boldsymbol{\sigma^{2}})$
   
   \STATE ${\bar{\mu}}^{(l)}, {\bar{\sigma}{{}^{2}}^{(l)}} = \hat{f}_c(\boldsymbol{\bar{\mu}}, \boldsymbol{{\sigma}^2})$
   
    \STATE // Compute marginal contribution of $i$ to coalitions of size $k$ 
   
   \STATE $\vec{R}^c[i] = \vec{R}^c[i] + \frac{1}{K} ({\bar{\mu}}^{(l)} - {\mu}^{(l)}) $
   \ENDFOR

   \ENDFOR
   \STATE {\bfseries Output:} Approximate Shapley values $\vec{R}^c$
\end{algorithmic}
\end{algorithm}

\begin{figure*}[t!]
    \centering
    \includegraphics[width=\textwidth]{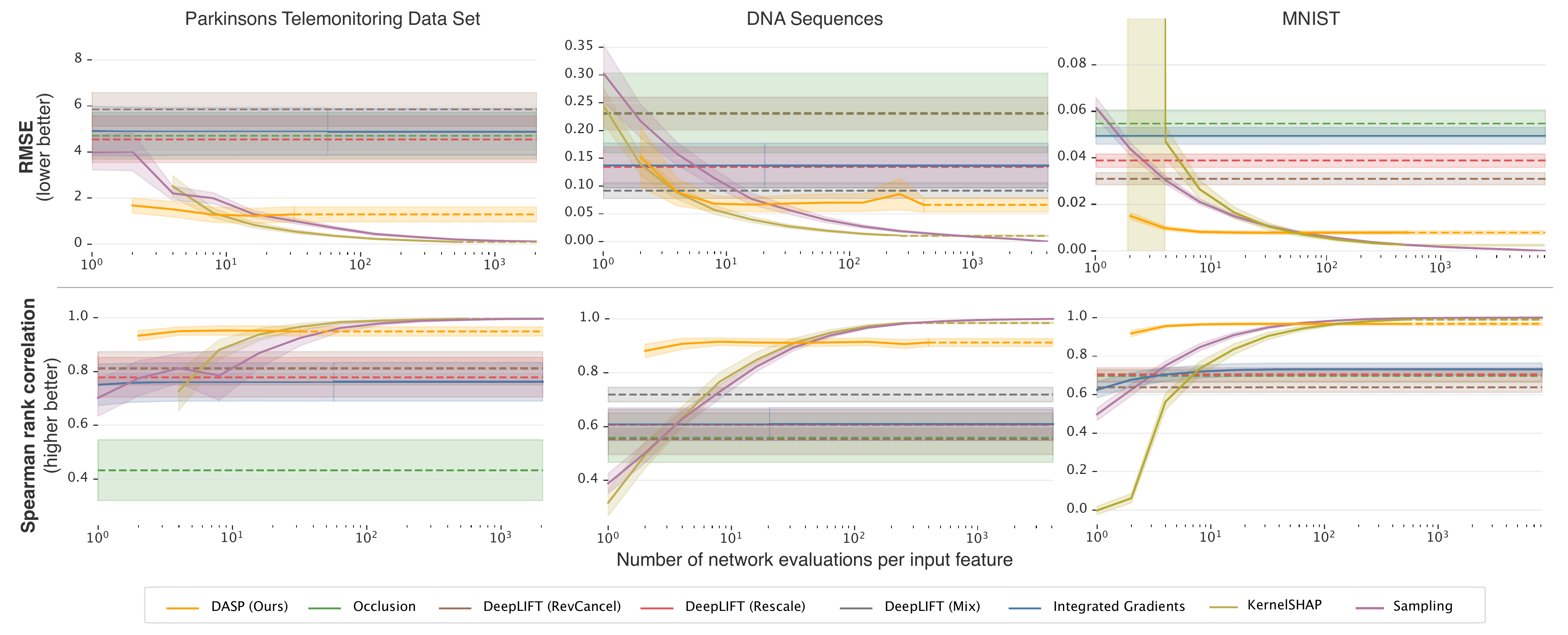}
    \vskip -0.15in
    \caption{Comparison of attribution methods on three datasets according to RMSE and Spearman correlation, with respect to the number of network evaluations. For each method, we report the mean and standard deviation computed over at least 50 samples on each dataset.  The ground truth is approximated using Shapley sampling for DNA Sequences and MNIST. DeepLIFT and Occlusion call for a fixed number of network evaluations. To enable a direct comparison, we project their performance on the $x$-axis using dashed lines.
    DASP evaluations are adjusted to reflect the doubled number of evaluations to propagate both mean and variance. KernelSHAP is run with no regularizer. 
    Vertical scales adjusted for the sake of readability. Best seen in electronic form.}
    \vspace{-0.10in}
    \label{fig:comparison}
\end{figure*}

Computing Deep Shapley values requires $N$ network evaluations for each input feature, if we test all possible coalition sizes (i.e. $k=0,...,N-1$). As a result, the algorithm approximates Shapley values for $N$ input features in $\mathcal{O}(N^2)$ network evaluations, compared to $\mathcal{O}(2^N)$ of the exact computation. 
% On the other hand, notice that our algorithm can be used to compute the Shapley values with respect to all hidden and output units \textit{at the same time}, while backpropagation-based methods would require to run the algorithm for each unit to be explained. This is useful to investigate what activates a specific hidden unit. 
%
In order to further reduce the computational cost, we can perform a secondary approximation by reducing the number of coalition sizes that we consider. If $K$ is the number of coalition sizes tested, we need $\mathcal{O}(KN)$ evaluations. In the next section, we show empirically that $K$ can be much lower than $N$. In order to ensure as much diversity as possible, we pick coalition sizes $k = [k_1,...,k_K]$ equally apart from each other.

\section{Experiments}

In this section, we report experiments running DASP\footnote{DASP implementation: http://bit.ly/DASPCode} alongside Integrated Gradients, DeepLIFT Rescale, DeepLIFT, RevealCancel, Occlusion, Shapley sampling and KernelSHAP using three datasets and different network architectures. Instead, we omit the comparison with DeepSHAP as this method is equivalent to DeepLIFT Rescale when used with a (fixed) zero baseline.
Our main goal is to demonstrate how DASP can be effectively used to approximate Shapley values and compare it with state-of-the-art attribution methods on this task.
Notice that Shapley sampling and KernelSHAP (without regularizer) are guaranteed to converge to exact Shapley values, given enough samples. This is not the case for DASP, which is not a sampling-based method. In this case, our goal is to show that sampling-based methods require significantly more model evaluations than DASP to reach the same approximation error.

All experiments are run in Keras \cite{chollet2015keras}. Details about the architectures used are in Appendix F of the supplementary material.

\subsection{Evaluation metrics}
When the size of the input allows it, we compute exact Shapley values using Eq.\ \ref{eq:shapley}, and use it as the ground-truth. When the exact computation is not feasible, we run Shapley sampling until convergence as an approximation of the ground-truth. Since Shapley sampling is an unbiased estimator, this is a faithful approximation.

For the quantitative comparison, we report both the root mean squared error (RMSE) and the Spearman rank correlation over several input samples extracted from each test set. While the RMSE is useful to quantify the absolute average error of each attribution score, the Spearman rank correlation is used as a metric to assess whether two methods agree on the ranking of features based on their impact on the model output.
When discussing performance, we always compare the attribution methods by the number of network evaluations they require, as the wall-clock time is influenced by the efficiency of the different implementations. However, we do take into account that our probabilistic layers require about twice the number of operations of a normal layer (to propagate mean and variance), therefore we double the number of evaluations for DASP in all our results.

\subsection{Parkinsons disability assessment}

As a first test, we trained a fully-connected \ac{DNN} on the Parkinsons Telemonitoring Data Set \cite{Tsanas2010}. The goal of this regression task is to predict the Parkinson’s disease rating scale (UPDRS) that reflects the presence and severity of symptoms starting from 18 input features: subject age, subject gender, and 16 biomedical voice measures. UPDRS spans the range 0-176, with 0 representing healthy state and 176 representing total disabilities.

\begin{figure}[h!]
    \centering
    \includegraphics[width=\columnwidth]{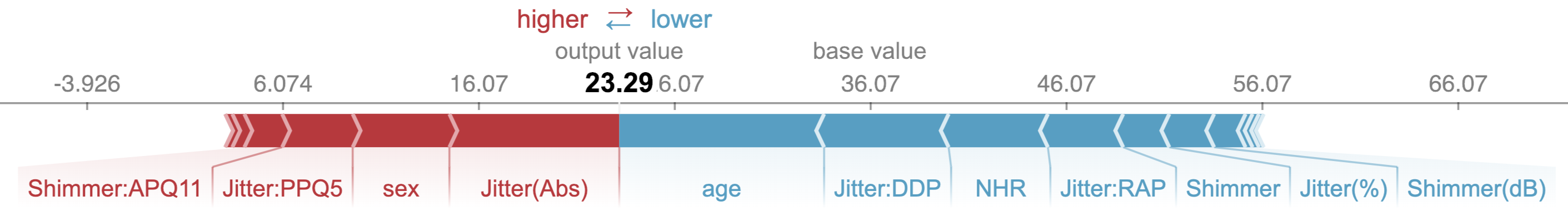}
    \vskip -0.1in
    \caption{Explanation for the UPDRS prediction ($=23.29$) of a single subject according to DASP. The explanation is displayed as force plot \cite{lundberg2018explainable}, where red(blue) indicates features that contributed making the score higher(lower). For example, \textit{sex} (male) causes the output score to increase while \textit{age} causes the output score to decrease. Best seen in electronic form.}
    \label{fig:forceplot}
\end{figure}

Fig.\ \ref{fig:forceplot} shows the explanation of the network prediction generated by our method for a single subject. An explanation, in this case, can tell the user which features the model has used for a certain prediction.
The natural question is whether the generated explanation is a reliable representation of the network behavior. Notice that the network behavior and the human intuition about how a certain task should be solved might not coincide in general (for example, the network might have learned patterns that the human expert was not aware of, or the network might mistakenly exploit correlations that have no causality). For this reason, we believe any qualitative comparison is dangerous. Instead, we have a clear ground truth against which we can evaluate our technique and others in this case. As the number of input features is limited to 18, we can compute the exact Shapley values for 100 samples from the test set, each requiring $2^{18}=262'144$ network evaluations. Fig.\ \ref{fig:comparison} shows that DASP produces a better approximation of Shapley values than other biased methods, even when a small number of coalition sizes is tested (i.e. $K \ll N$). Unbiased, sampling-based methods, instead, require significantly more evaluations before outperforming DASP.

\subsection{Classifying regulatory DNA sequences }
To test if the result holds on different architectures, we consider a classification task over DNA sequence inputs. A DNA sequence can be seen as a string over the alphabet A,C,G,T. We are interested in detecting short patterns (eg. GATAA or GATTA) within these sequences by employing a neural network with two 1D convolutional layers, global average pooling, and one fully-connected layer. This setup, which also includes some synthetically generated DNA sequences, was previously proposed as a benchmark for attribution methods \cite{shrikumar17a}. Since the DNA sequences have a length of 200 tokens, it is prohibitive to compute exact Shapley values. To approximate the ground-truth and enable a direct comparison of all methods against it, we run about 800'000 iterations of Shapley sampling on each of the 50 input samples until convergence. This process took about 45 minutes on our machine.

The results are reported in Fig.\ \ref{fig:comparison}. The original authors of the experiment suggest a specific combination of DeepLIFT propagation rules (Rescale for the two convolutional layers and RevealCancel for the dense layer) to obtain the best results on this dataset. This combination, identified as DeepLIFT (Mix) in Fig.\ \ref{fig:comparison}, also gives the best results among backpropagation-based methods in our experiment, outperformed only by DASP and sampling methods. On the other hand, KernelSHAP requires a significantly more evaluations than DASP to reach the same rank correlation.

\subsection{Digits classification (MNIST)}

\begin{figure}[h]
\begin{center}
\centerline{\includegraphics[width=\columnwidth]{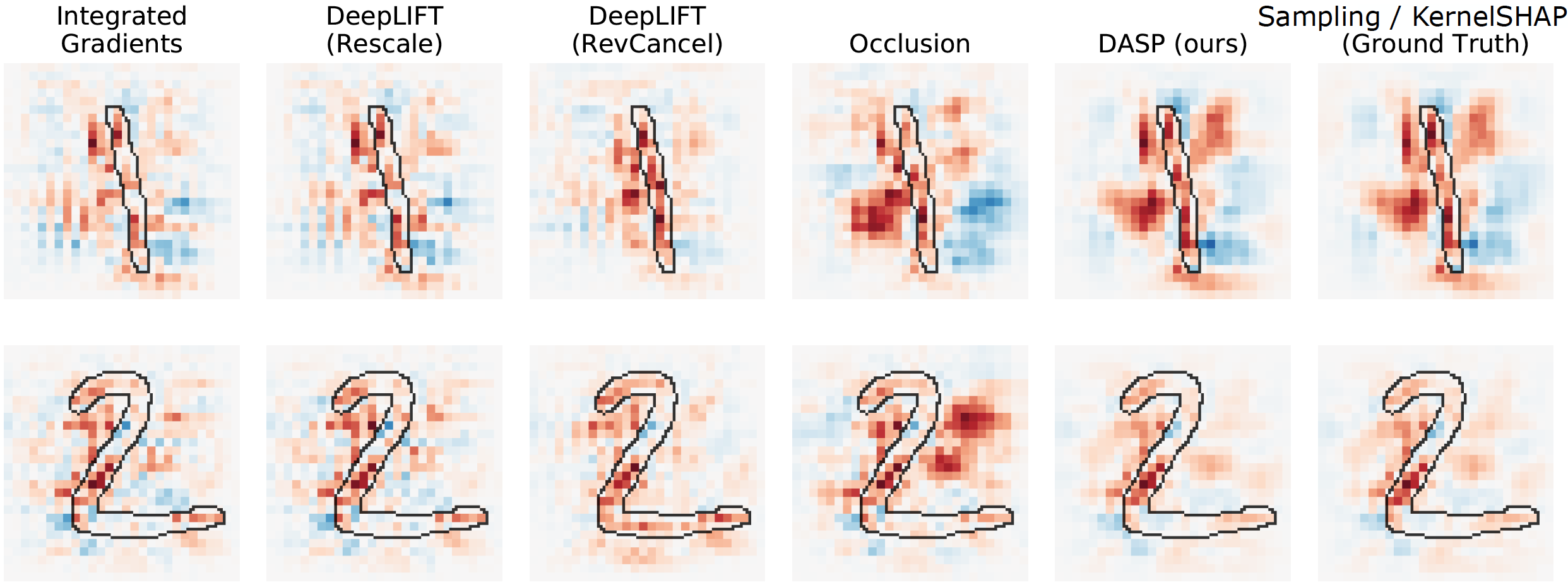}}
\vskip -0.1in
\caption{Attributions maps produced by different methods on MNIST images. Red(blue) color indicates features (pixels) that positively(negatively) impact the network output score. KernelSHAP (not reported) converges to the same result of Sampling.}
\label{fig:mnist_maps}
\end{center}
\vskip -0.3in
\end{figure}

Finally, we train a convolutional neural network on MNIST \cite{lecun1998} to perform digit classification. We use the LeNet-5
architecture \cite{lecun1998}, which consists of two convolutional layers, each followed by a max-pooling layer, and three final dense layers.  Again, we use Shapley sampling to approximate the ground-truth for the 784 features (pixels) of 50 test images, which took about 4 hours for 6M evaluations on our machine.

Fig.\ \ref{fig:mnist_maps} shows an example of the resulting attribution maps. Notice that backpropagation-based methods tend to produce more noisy explanations than perturbation-based methods. We speculate that this happens because backpropagation-based methods are more sensitive to local (and noisy) gradient information. Furthermore, Occlusion seems to under-estimate or over-estimate the importance of some areas of the image, as compared to DASP. These are reflected in the quantitative comparison in Fig.\ \ref{fig:comparison}. As in the previous experiments, we see a significant gap between the accuracy of DASP and other biased approximators, as well as a significant gap in the number of samples required by Shapley sampling and KernelSHAP to reach DASP performance. Notice that, in this case, KernelSHAP performs worse than simple sampling. We speculate this happens because of the linearity assumption of KernelSHAP, affecting more the more complex models.

\vskip -1cm
\section{Conclusions}
In this work, we discussed Shapley values as an attribution method for \acp{DNN}. First, we showed that several existing attribution methods reduce to computing Shapley values when applied to linear models. On the other hand, when the model is non-linear, Shapley values remain the only method that satisfies a number of desirable theoretical properties, which strongly motivates their use for reliable explanations.

As computing Shapley values is often unfeasible, we then proposed  Deep Approximate Shapley Propagation (DASP), a novel perturbation-based method that approximates Shapley values using uncertainty propagation in \acp{DNN}. DASP requires a polynomial number of network evaluations. While it is not guarantee to recover exact Shapley values, we showed empirically that other sampling-based methods require significantly more evaluations to achieve the same approximation error.
We have also shown that DASP outperforms  state-of-the-art backpropagation-based methods, which are fast but coarse Shapley values approximators. 

DASP can be considered a novel application for the field of uncertainty propagation in \acp{DNN}, although we would like to remark how it is not constrained to LPN for this purpose. As research on this area continues, we expect to be able to extend DASP for recurrent neural networks, and we hope that new probabilistic frameworks will enable the derivation of theoretical guarantees and even better approximations.

\clearpage
\FloatBarrier

% \section*{External resources}
% The DASP source code and the supplementary material are available online: http://bit.ly/DASPCode.

\bibliography{paper}
\bibliographystyle{icml2019}

\end{document}